\documentclass[11pt,a4paper]{article}
\usepackage[hyperref]{eacl2021}
\usepackage{times}
\usepackage{latexsym}
\usepackage{booktabs}
\usepackage{tabularx}
\usepackage{graphicx}
\usepackage{amsmath}
\usepackage{amsfonts}
\usepackage{amssymb}

\usepackage{booktabs}

\usepackage{microtype}

\aclfinalcopy 
\def\aclpaperid{678} 

\title{Cross-lingual Contextualized Topic Models with Zero-shot Learning}

\author{Federico Bianchi \\
    Bocconi University \\ 
    Via Sarfatti 25, 20136 \\ Milan, Italy \\  f.bianchi@unibocconi.it \\\And
  Silvia Terragni \\
    University of Milano-Bicocca \\
    Viale Sarca 336, 20126 \\
    Milan, Italy  \\
       s.terragni4@campus.unimib.it\\ \And
  Dirk Hovy \\
    Bocconi University \\ 
    Via Sarfatti 25, 20136 \\ Milan, Italy \\
    dirk.hovy@unibocconi.it \\ \AND
     Debora Nozza \\
    Bocconi University \\ 
    Via Sarfatti 25, 20136 \\ Milan, Italy \\
    debora.nozza@unibocconi.it \\ \And
     Elisabetta Fersini \\
    University of Milano-Bicocca \\ 
    Viale Sarca 336, 20126 \\ Milan, Italy \\
    elisabetta.fersini@unimib.it}
\date{}

\begin{document}

\setcounter{topnumber}{8}
\setcounter{bottomnumber}{8}
\setcounter{totalnumber}{8}

\maketitle
\begin{abstract}
Many data sets (e.g., reviews, forums, news, etc.) exist parallelly in multiple languages. They all cover the same content, but the linguistic differences make it impossible to use traditional, bag-of-word-based topic models. Models have to be either single-language or suffer from a huge, but extremely sparse vocabulary. Both issues can be addressed by transfer learning.
In this paper, we introduce a zero-shot cross-lingual topic model. Our model learns topics on one language (here, English), and predicts them for unseen documents in different languages (here, Italian, French, German, and Portuguese). We evaluate the quality of the topic predictions for the same document in different languages. Our results show that the transferred topics are coherent and stable across languages, which suggests exciting future research directions.
\end{abstract}

\section{Introduction}
Topic models~\cite{Blei03lda, blei2012probabilistic} allow us to find the main themes and overarching tropes in textual data. However, traditional methods are language-specific and cannot be used in a \textit{transferable manner}. They rely on a fixed vocabulary specific to the training language. 

Therefore, currently available topic models suffer from two limitations: (i) they cannot handle unknown words by default, and (ii) they cannot easily be applied to other languages - except the one in the training data - since the vocabulary would not match. Training on several languages together, though, results in a vocabulary so vast that it creates problems with parameter size, search, and overfitting~\cite{boyd2014care}. Traditional topic modeling provides methods to extract meaningful word distributions from ``unstructured'' text but requires language-specific bag-of-words (BoW) representations~\cite{Boyd-GraberB09multilingual,JagarlamudiD10multilingual}. 


A cross-lingual setup proves ideal for \textit{transfer learning}: provided that the gist of topics is the same across languages, we can learn this gist on texts in one language and then apply it to others. This setup is \textit{zero-shot learning}: we train a model on one language and test it on several other languages to which the model had no access during training.

To this end, we need to leverage external information to support the topic modeling task. Indeed, topic models have often gained significant advantages from introducing external knowledge, e.g., document relationships~\cite{yang2015birds,wang2020deep,terragni2020constrained,Terragni20negativeresults} and word embeddings ~\cite{nozza2016unsupervised,Li16,zhao2017metalda,dieng2019diversity}.  Recently, pre-trained contextualized embeddings, e.g., BERT~\cite{devlin2018bert} embeddings, have enabled exciting new results in several NLP tasks~\cite{rogers2020primer,nozza2020mask}. More importantly, there do exist contextualized embeddings that are also multilingual.

This paper introduces a novel neural topic modeling architecture in which we \textit{replace} the input BoW document representations with multilingual \textit{contextualized} embeddings. Neural topic models take in input the document BoW representations, which provide valuable symbolic information; however, this information's structure is lost after the first hidden layer in any neural architecture. We, therefore, hypothesize that contextual information can replace the BoW representation. 

We use a neural encoding layer for the pre-trained document representations from a contextualized embedding model input (e.g., BERT) before the neural topic model's sampling process. This change allows us to address the two limitations mentioned above jointly: (i) our approach solves the problem of dealing with unseen words at test time since we do not need them to have a BoW representation; moreover, (ii) the model infers topics on unseen documents in languages other than the one in the training data. The inferred topics consist of tokens from the training language and can be applied to any supported test language. We show the high quality of the resulting topics for four test languages both quantitatively and qualitatively. 

To the best of our knowledge, there is no prior work on zero-shot cross-lingual topic modeling. Our model can be applied to new languages after training is complete and does not require external resources, alignment, or other conditions. Nonetheless, the flexibility of the input means our model will benefit from any future improvement of language modeling techniques.


\paragraph{Contributions} We release a novel neural topic model that relies on language-independent representations to generate topic distributions. We show that this input can replace the standard input BoW without loss of quality. We show that its multilingual representations enable zero-shot cross-lingual tasks. The solution we propose is straightforward and does not require high computational resources since it can efficiently run on common laptops (see Appendix). 
We have implemented the tool as a documented python package available at \url{https://github.com/MilaNLProc/contextualized-topic-models}.

\section{Contextualized Neural Topic Models}
We extend Neural-ProdLDA~\cite{srivastava2017avitm}, one of the most recent and promising approaches of neural topic modeling, based on the Variational AutoEncoder (VAE)~\cite{Kingma13variationalbayes}. The neural variational framework trains an inference network, i.e., a neural network that directly maps the BoW representation of a document onto a continuous latent representation. A decoder network then reconstructs the BoW by generating its words from the latent document representation. This latent representation is sampled from a Gaussian distribution parameterized by $\mu$ and $\sigma^2$ that are part of the variational inference framework~\cite{Kingma13variationalbayes} --- see ~\cite{srivastava2017avitm} for more details. 

\begin{figure}[ht]
    \centering
    \includegraphics[width=0.7\columnwidth]{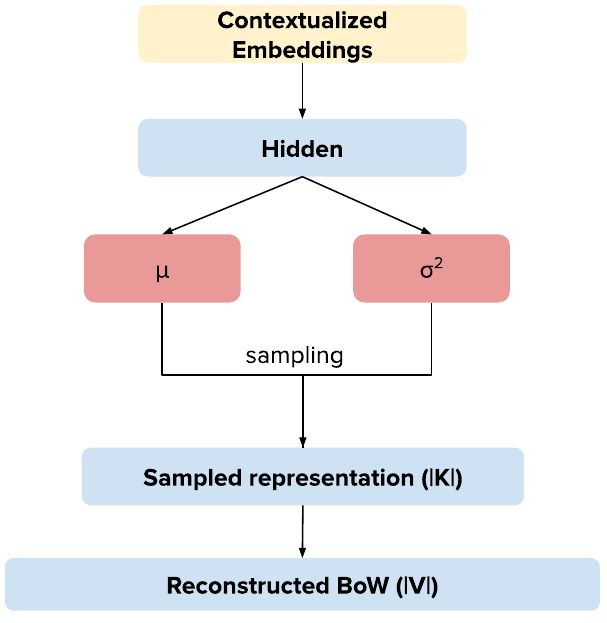}
    \caption{High-level schema of the architecture for the proposed contextualized neural topic model.}
    \label{fig:multi:topic}
\end{figure}

We replace the input BoW in Neural-ProdLDA with pre-trained multilingual representations from SBERT~\cite{reimers2019sentence}, a recent and effective model for contextualized representations. In Figure~\ref{fig:multi:topic}, we sketch the architecture of our contextualized neural topic model. The final \textit{reconstructed BoW} layer is still a component of our model: the BoW representation is necessary for the model's training to obtain the topic indicators (i.e., the most likely words representing a topic), but it becomes useless during testing. 

{}

Our proposed model, \textbf{Zero-Shot Topic Model} (ZeroShotTM), is trained with input document representations that account for word-order and contextual information, overcoming one of the central limitations of BoW models. 
Moreover, the use of language-independent document representations allows us to do zero-shot topic modeling for unseen languages. This property is essential in low-resource settings in which there is little data available for the new languages. Because multilingual contextualized representations exist for multiple languages, it allows zero-shot modeling in a \textit{cross-lingual} scenario. Indeed, ZeroShotTM is language-independent: given a contextualized representation of a new language as input,\footnote{As long as a multilingual model - like multilingual BERT - covers it.} it can predict the topic distribution of the document. The predicted topic descriptors, though, will be from the training language. 
Let us also notice that our method is agnostic about the choice of the neural topic model architecture (here, Neural-ProdLDA), as long as it extends a Variational Autoencoder.

\section{Experiments}
Our experiments evaluate two main hypotheses: (i) \textit{we can define a topic model that does not rely on the BoW input but instead uses contextual information;} (ii) \textit{the model can tackle zero-shot cross-lingual topic modeling}. The Appendix contains more details about the experiments (e.g., code, data, runtime, replication details).

\paragraph{Datasets} We use datasets collected from English Wikipedia abstracts from DBpedia.\footnote{\url{https://wiki.dbpedia.org/downloads-2016-10}} The first dataset (W1) contains 20,000 randomly sampled abstracts. The second dataset (W2) contains 100,000 English documents. We use 99,700 documents as training and consider the remaining 300 documents as the test set. We collect the 300 respective instances in Portuguese, Italian, French, and German. This collection creates a test set of comparable documents, i.e., documents that refer to the same entity in Wikipedia, but in different languages. 

We extract only the first 200 tokens of each abstract to reduce the length limit's effects in the tokenization process. In particular, we use the efficient and effective SBERT~\cite{reimers2019sentence},\footnote{\url{https://github.com/UKPLab/sentence-transformers}} using the multilingual model,\footnote{We use the \textit{distiluse-base-multilingual-cased} embeddings for this experiment available on the authors' repository.} on this unpreprocessed text. We then remove stopwords and use the most frequent remaining 2,000 words to create the English vocabulary for BoW model comparisons.

\subsection{To Contextualize or Not To Contextualize}

First, we want to check if \textbf{ZeroShotTM} maintains comparable performance to other topic models; if this is true, we can then explore its performance in a cross-lingual setting. Since we use only English text, in this setting we use English representations.\footnote{We use the \textit{bert-base-nli-mean-tokens} model.} 

\begin{table}[ht]
    \centering
    \begin{tabular}{ccc} \toprule
        \textbf{Model} & $\tau$ (50) & $\tau$ (100)  \\ \midrule 
        ZeroShotTM & 0.1632 & 0.1381  \\
        Combined TM & 0.1644 & \textbf{0.1409*} \\
        Neural-ProdLDA  & \textbf{0.1658} & 0.1285  \\
        LDA & -0.0246 & -0.0757 \\ \bottomrule
    \end{tabular}
    \caption{NPMI Coherences on W1 data set. * denotes the statistically significant results (t-test).}
    \label{tab:to_con_or_not}
\end{table}{}

We compare \textbf{ZeroShotTM} on W1 with: (i) Combined TM~\cite{bianchi2020pretraining}, an extension of Neural-ProdLDA that concatenates \textit{both} BoWs and SBERT representations (transformed to the same dimension of the BoWs) as inputs to the model, (ii) Neural-ProdLDA~\cite{srivastava2017avitm}, and (iii) LDA~\cite{Blei03lda}. 

We compute the topic coherence~\cite{Lau14npmi} via NPMI ($\tau$) for 50 and 100 topics averaging models' results over 30 runs.
We report the results in Table~\ref{tab:to_con_or_not}. ZeroShotTM obtains comparable results to Combined TM and Neural-ProdLDA in this setting. Contextualized embeddings \textit{can} replace BoW input representations without loss of coherence.

\subsection{Zero-shot Cross-Lingual Topic Modeling}
\textbf{ZeroShotTM} can be used for zero-shot cross-lingual topic modeling. We evaluate multilingual topic predictions on the multilingual abstracts in W2. We use SBERT~\footnote{\url{https://github.com/UKPLab/sentence-transformers}} to generate multilingual embeddings as the input of the model.

\begin{table*}[h]
    \centering
    \begin{tabular}{l|rrr|rrr} \toprule
        \textbf{Lang} & \textbf{Mat25}$\uparrow$  & \textbf{KL25}$\downarrow$ & \textbf{CD25}$\uparrow$ & \textbf{Mat50}$\uparrow$ & \textbf{KL50}$\downarrow$ & \textbf{CD50}$\uparrow$ \\ \midrule
        IT & 75.67& 0.16 & 0.84 &62.00    & 0.21& 0.75 \\
        FR & 79.00 & 0.14 & 0.86 & 63.33    & 0.19 &0.77\\
        PT     &  78.00 &0.14 & 0.85 & 68.00 &0.19 & 0.79\\
        DE &    79.33 & 0.15 & 0.85    & 64.33 & 0.20 & 0.77\\ \midrule
        ZeroShotTM Avg &  \textbf{78.00} & \textbf{0.15} & \textbf{0.85}    & 64.41 & 0.20 & 0.77\\
        \midrule  
        Ori Avg & 76.00  & \textbf{0.15} & 0.84 & \textbf{69.00} & \textbf{0.19} & \textbf{0.79} \\
        Uni & 4.00  &0.75 & --- & 2.00 & 0.85 & --- \\ \bottomrule
    \end{tabular}
    \caption{Match, KL, and centroid similarity for 25 and 50 topics on various languages on W2.}
    \label{tab:results:multi}
\end{table*}{}

\subsubsection{Quantitative Evaluation}

Since the predicted document-topic distribution is subject to a stochastic sampling process, we average it over 100 samples to obtain a better estimate.

\paragraph{Metrics}

We expect the topic distributions over a set of comparable documents (e.g., in English and Portuguese) to be similar to each other.
We compare the topic distributions of each abstract in a test language with the topic distribution of the respective abstract in English, which is the training language. Note that the English test document is also unseen, i.e., the training data does not include it.
We evaluate our model on three different metrics. 
The first metric is \textbf{matches}, i.e., the percentage of times the predicted topic for the non-English test document is the same as for the respective test document in English. The higher the scores, the better.

To also account for similar but not exactly equal topic predictions, we compute the \textbf{centroid embeddings} of the five words describing the predicted topic for both English and non-English documents. Then we compute the cosine similarity between those two centroids~(CD). 

Finally, to capture the \textbf{distributional similarity}, we also compute the KL divergence between the predicted topic distribution on the test document and the same test document in English. Here, lower scores are better, indicating that the distributions do not differ by much.

\paragraph{Automatic Evaluation}

We use two baselines: the first one (Ori) consists of performing topic modeling on documents translated into English via DeepL.\footnote{\url{https://www.deepl.com/}} While this is an easily accessible baseline, automatic translation is costly and may introduce bias in the representations (as shown by~\newcite{hovy2020you}). We compare the predicted topics of each translated document to the ones predicted for the original English document (as done above).  
The second baseline is a uniform distribution (Uni): we compute all the metrics over a uniform distribution (this baseline gives a lower bound). 

Table~\ref{tab:results:multi} shows the evaluation results of our model in the zero-shot context. Note that because we trained on English data, the topic descriptors are in English. Topic predictions are significantly better than the uniform baselines: more than 70\% of the times, the predicted topic on the test set matches the topic of the same document in English. The CD similarity suggests that even when there is no match, the predicted topic on the unseen language is at least similar to the one on the English testing data. Simultaneously, the predictions for the contextualized model are in line with the ones obtained using the translations (Ori Avg), showing that our model is capable of finding good topics for documents in unseen languages without the need for translation.

\paragraph{Manual Evaluation} We rated the predicted topics for 300 test documents in five languages (thus, 1500 docs including English) on an ordinal scale from 0-3. A 0 rate means that the predicted topic is wrong, a 1 rate means the topic is somewhat related, a 2 rate means the topic is good, and a 3 rate means the topic is entirely associated with the considered document. 
Table~\ref{tab:manual:evaluation} shows the results per language. We evaluate the inter-rater reliability using Gwet AC1 with ordinal weighting~\cite{gwet2014handbook}. The resulting value of 0.88 indicates consistent scoring.

\begin{table}[h]
    \centering
    \begin{tabular}{c|c} \toprule
        \textbf{Language} & \textbf{Average Topic Quality}  \\ \midrule
        English & 2.35              \\
        Italian &  2.29 \\
        French &  2.22 \\
        Portuguese &  2.26 \\
        German & 2.19 \\ \midrule
        Average & 2.26 \\ \bottomrule
    \end{tabular}
    \caption{Average topic quality (out of 3).}
    \label{tab:manual:evaluation}
\end{table}{}

\begin{table*}[ht]
\small
    \centering
    \begin{tabular}{l|l|l}
        \normalsize\textbf{Lang} & \normalsize\textbf{Sentence} & \normalsize\textbf{Predicted Topic} \\ \toprule
         EN & Blackmore's Night is a British/American traditional folk rock duo [...] &  rock, band, bass, formed \\
         IT & I Blackmore's Night sono la band fondatrice del renaissance rock [...]  & rock, band, bass, formed \\
         PT & Blackmore's Night é uma banda de folk rock de estilo renascentista [...] &  rock, band, bass, formed \\ 
         \midrule
         EN & Langton's ant is a two-dimensional Turing machine with [...] & mathematics, theory, space, numbers \\
         FR & On nomme fourmi de Langton un automate cellulaire  [...] & mathematics, theory, space, numbers\\
         DE & Die Ameise ist eine Turingmaschine mit einem zweidimensionalen [...] & mathematics, theory, space, numbers  \\
         \midrule
         EN & The Journal of Organic Chemistry, colloquially known as JOC or [...] & journal, published, articles, editor \\
         IT & Journal of Organic Chemistry è una rivista accademica [...] &  journal, published, articles, editor \\
         PT & Journal of Organic Chemistry é uma publicação científica [...] & journal, published, articles, editor \\ \midrule
         EN & Joan Brossa [...] was a Catalan poet, playwright, graphic designer  [...] & book, french, novel, written \\
         IT & 
         Fu l'ispiratore e uno dei fondatori della rivista "Dau al Set"[...]
         & book, french, novel, written \\
         PT & Joan Brossa i Cuervo [...] foi um poeta, dramaturgo, artista plástico [...] &painting, art, painter, works \\
         \bottomrule
    \end{tabular}
    \normalsize
    \caption{\normalsize Examples of zero-shot cross-lingual topic classification in various languages with ZeroShotTM.}
    \label{tab:examples:languages}
\end{table*}

\subsubsection{Qualitative Evaluation}

In Table~\ref{tab:examples:languages}, we show some examples of topic predictions on test languages. Our model predicts the main topic for all languages, even though they were unseen during training.

The predicted topic is generally consistent with the text. I.e., the topics are easily interpretable and give the user a coherent impression. In some circumstances, noise biases the results: dates in the abstract tend to make the model predict a topic about time. Another interesting case is the abstract of the artist Joan Brossa, who was both a poet and a graphic designer. In the English and Italian abstract, the model has discovered a topic related to writing. In constrast, in the Portuguese abstract, the model has found a topic related to art, which is still meaningful.

\section{Related Work}
While not in a zero-shot fashion, several researchers have studied multilingual and cross-lingual topic modeling ~\cite{MaN17multil,GutierrezSLMG16multi,HaoP18multil, HeymanVM16multil, LiuDM15multil, KrstovskiSK16multil}. 

The first model proposed to process multilingual corpora with LDA is the Polylingual Topic Model by~\newcite{MimnoWNSM09multil}. It uses LDA to extract language-consistent topics from parallel multilingual corpora, assuming that translations share the same topic distributions. Models that transfer knowledge on the document level have many variants, including \cite{HaoP18multil, HeymanVM16multil, LiuDM15multil, KrstovskiSK16multil}.
However, existing models require to be trained on multilingual corpora and are always language-dependent: they cannot predict the main topics of a document in an unseen language. 

Other models use multilingual dictionaries \cite{Boyd-GraberB09multilingual, JagarlamudiD10multilingual}, requiring some predefined mapping.
Embeddings, both for words and documents, have been shown to capture a wide range of semantic, syntactic, and social aspects of language \cite{hovy-purschke-2018-capturing,rogers2020primer}. Our work adds language-independent topics to that list.

\section{Conclusions}
We propose a novel neural architecture for cross-lingual topic modeling using contextualized document embeddings as input. Our results show that (i) contextualized embeddings can replace the input BoW representations and (ii) using contextualized representations allows us to tackle zero-shot cross-lingual topic modeling. The resulting model can be trained on any one language and applied to any other language for which embeddings are available. 

\section{Acknowledgements}
Federico Bianchi, Dirk Hovy, and Debora Nozza are members of the Bocconi Institute for Data Science and Analytics (BIDSA) and the Data and Marketing Insights (DMI) unit.

\bibliography{eacl2021}
\bibliographystyle{acl_natbib}

\appendix

\section{Datasets}

We used the English DBpedia 2016-10 abstract dump\footnote{\url{https://wiki.dbpedia.org/downloads-2016-10}} to create our datasets. 

\paragraph{\textbf{W1}} We randomly sampled 20,000 documents from the English DBpedia abstract dump to create our first set of documents. We created W1 to provide a quick collection of documents to test if our Contextual TM performance does not decrease significantly.

\paragraph{\textbf{W2}}  We collected 100,000 abstracts sampling randomly from those that had at least 200 chars. Given this set, we extracted 300 random English abstracts. Given the random abstracts, we retrieved the respective version in other languages using the DBpedia SPARQL endpoint.\footnote{\url{https://dbpedia.org/sparql}} We manually evaluated the quality of the 300 abstracts since we looked at each of those during our manual evaluation, finding no mismatch between the abstract and no corrupted text.

\subsection{Preprocessing}

We followed a standard pre-processing pipeline to generate the preprocessed set of documents. We removed punctuation, digits, and nltk's English stop-words.\footnote{\url{https://www.nltk.org/}}
Following other researchers, we selected 2,000 as the maximum number of words for the BoW, and thus we kept in the abstracts only the 2,000 most frequent words. 

\section{Models and Baselines}

\subsection{Neural-ProdLDA} We use the implementation made available by~\newcite{Carrow2018avitm} since it is the most recent and with the most updated packages (e.g., one of the latest versions of PyTorch). The model is trained for 100 epochs. We use ADAM optimizer (with a learning rate equal to 2e-3). The inference network is composed of a single hidden layer and 100-dimension of softplus units. The priors over the topic and document distributions are learnable parameters. Momentum is set to 0.99, the learning rate is set to 0.002, and we apply 20\% of drop-out to the hidden document representation. The batch size is equal to 200. More details related to the architecture can be found in the original work~\cite{srivastava2017avitm}.

\subsection{ZeroShot TM} The model and the hyper-parameters are the same for Neural-ProdLDA, with the difference that we replace the BoW with SBERT features. The model is trained for 100 epochs. We use ADAM optimizer. 

\subsection{Combined TM} The model~\cite{bianchi2020pretraining}\footnote{\url{https://github.com/MilaNLProc/contextualized-topic-models}} and the hyper-parameters are the same used for Neural-ProdLDA with the difference that we also use SBERT features in combination with the BoW: we take the SBERT embeddings, apply a (learnable) function/dense layer $R^{512} \rightarrow R^{|V|}$ and concatenate the representation to the BoW. The model is trained for 100 epochs. We use ADAM optimizer.

\subsection{LDA} We use Gensim's\footnote{\url{https://radimrehurek.com/gensim/models/ldamodel.html}} implementation of this model. The hyper-parameters $alpha$ and $beta$, controlling the document-topic and word-topic distribution respectively, are estimated from the data during training.

\section{Computing Infrastructure}
We ran experiments on two common laptops, equipped with a GeForce GTX 1050 (running CUDA 10). As our experiments show, the models can be easily run with basic hardware (having a GPU is better than just using CPU, but the experiments can also be replicated on CPU). Both laptops have 16GB of RAM.

\subsection{Runtime}
Our implementation is written in PyTorch and runs on both GPU and CPU. Table~\ref{tab:my_label} shows the runtime for one epoch of both our Combined TM and Neural-ProdLDA for 25 and 50 topics on the GeForce GTX 1050. Neural-ProdLDA is slightly faster than our ZeroShotTM. This is due to the additional representation that cannot be encoded as a sparse matrix. However, we believe that these numbers are comparable and make our model easy to use even with common hardware.

\begin{table}[h]
    \centering
    \begin{tabular}{c|cc} \toprule
       Model  & W1 (25) & W1 (50)  \\ \midrule
      \textbf{ZeroShot TM} & 1.2s & 1.2s \\
       \textbf{Neural-ProdLDA} & 0.8s & 0.9s \\ \bottomrule
    \end{tabular}
    \caption{Time to complete one epoch on the \textbf{W1} dataset with 25 and 50 topics.}
    \label{tab:my_label}
\end{table}

\section{Source Code}
\subsection{Development}
Our software is available as a Python package that a user can easily install.\footnote{\url{https://github.com/MilaNLProc/contextualized-topic-models}}

\end{document}